\documentclass{article}
\usepackage{spconf,amsmath,graphicx}
\include{pythonlisting}
\usepackage{algorithm}
\usepackage[noend]{algpseudocode}
\usepackage{slashbox}
\usepackage{multirow}
\usepackage{url}
\usepackage{cite}

\title{A network of deep neural networks for distant speech recognition}
\name{Mirco Ravanelli$^{\S,\star}$\thanks{$^\star$This work was done while the author was visiting the Montreal Institute for Learning Algorithms (MILA)  and was supported by the FBK mobility programme.}, Philemon Brakel$^{\dagger}$, Maurizio Omologo$^{\S}$, Yoshua Bengio$^{\dagger}$}

\address{$^{\S}$Fondazione Bruno Kessler, Trento, Italy \\ 
$^{\dagger}$Universit\'e de Montr\'eal, Montr\'eal, Canada} 
\begin{document}
\ninept
\maketitle
\begin{abstract}
Despite the remarkable progress recently made in distant speech recognition, state-of-the-art technology still suffers from a lack of robustness, especially when adverse acoustic conditions characterized by non-stationary noises and reverberation are met.

A prominent limitation of current systems lies in the lack of matching and communication between the various technologies involved in the distant speech recognition process. The speech enhancement and speech recognition modules are, for instance, often trained independently. Moreover, the speech enhancement normally helps the speech recognizer, but the output of the latter is not commonly used, in turn, to improve the speech enhancement.

To address both concerns, we propose a novel architecture based on a network of deep neural networks, where all the components are jointly trained and better cooperate with each other thanks to a full communication scheme between them. Experiments, conducted using different datasets, tasks and acoustic conditions, revealed that the proposed framework can overtake other competitive solutions, including recent joint training approaches.
\end{abstract}
\begin{keywords}
speech recognition, speech enhancement, joint training, deep neural networks
\end{keywords}
\section{Introduction}
\label{sec:intro}
Distant Speech Recognition (DSR) is a technology of fundamental importance towards more flexible and effective human-machine interfaces. Such a technology might eventually allow users to access speech recognition services even when adverse acoustic conditions are met or when a distant-talking (far-field) interaction with a machine is required. A crucial role in improving current solutions is being played by deep learning \cite{Goodfellow-et-al-2016-Book}, which has recently contributed to outperforming previous HMM-GMM speech recognizers \cite{lideng}.
The progress in the field was also fostered by the considerable success of some international challenges such as CHiME \cite{chime3} and REVERB \cite{revch_short}.

Despite the great efforts of the past years, state-of-the-art techniques still exhibit a significant lack of robustness to acoustic conditions characterized by non-stationary noises and acoustic reverberation \cite{adverse}. 
To counteract such adversities, most DSR systems  
must rely on a combination of several interconnected technologies \cite{nakatani_short}, including methods for speech enhancement \cite{BrandWard}, speech separation \cite{bss}, acoustic event classification \cite{aed1,aed5}, speaker identification \cite{Beigi}, just to name a few.

A significant limitation of most current techniques lies in the lack of matching between the various modules being combined.
For example, speech enhancement and speech recognition are often designed independently and, in several cases, the enhancement part is tuned according to metrics which are not directly correlated with the final speech recognition performance. 
Another potential limitation is the lack of communication between the various modules of the DSR system. Most systems are based on a unidirectional information flow across the DSR pipeline. Speech enhancement tries to help the speech recognizer, but the latter does not contribute, in turn, to improve the performance of the speech enhancement module.
We argue that establishing this missing link can nevertheless be very useful, since a hint on the recognized phone sequence might help the speech enhancement in performing its task. 
A fruitful integration between the various systems, however, was very difficult for many years, mainly due to the different nature of the technologies involved at the various steps.  
Nevertheless, the recent success of deep learning has not only largely contributed to the substantial improvement of the speech recognition part of a DSR system \cite{pawel2,hain,ravanelli15}, but has also enabled the development of competitive DNN-based speech enhancement solutions \cite{dnn_se2,dnn_se3}, making an effective integration between such modules easier. 

Within the DNN framework, the adoption of a joint training approach between speech enhancement and speech recognition DNNs has recently been proposed to mitigate the lack-of-matching issue \cite{joint2,joint3,joint6,joint7,joint4,joint5,joint_icassp2016,ravanelli_SLT}. 
The core idea is to pipeline such DNNs and to jointly update their parameters as if they were within a single bigger network. 
Despite the considerable effectiveness of joint training, most of these methods suffer from a lack of communication, since the standard DSR pipeline based on a unidirectional communication between the speech enhancement and speech recognition is still employed.

In this paper, we attempt to evolve standard joint training approaches, by proposing a novel DNN architecture, potentially able to address at the same time both the lack-of-matching and the lack-of-communication arising in current systems.    
Such an architecture replaces the standard DSR pipeline with a \textit{network of deep neural networks} in which all the modules are jointly trained and better cooperate with each other thanks to a full communication between them. The architecture, after being properly unrolled, is trained with a variation of the standard back-propagation algorithm, which is based on a back-propagation of both speech recognition and speech enhancement gradients through the network of DNNs. 

The preliminary results reported in this paper confirm the effectiveness of this approach, showing that the proposed paradigm is able to overtake even recently proposed joint training methods.
The experimental validation has been carried out in a distant-talking scenario considering different training datasets, tasks and acoustic conditions.

\section{A NETWORK OF DEEP NEURAL NETWORKS}
\label{sec:ndnn}
\begin{figure}[t!]
\centering
\includegraphics[width=0.50\textwidth]{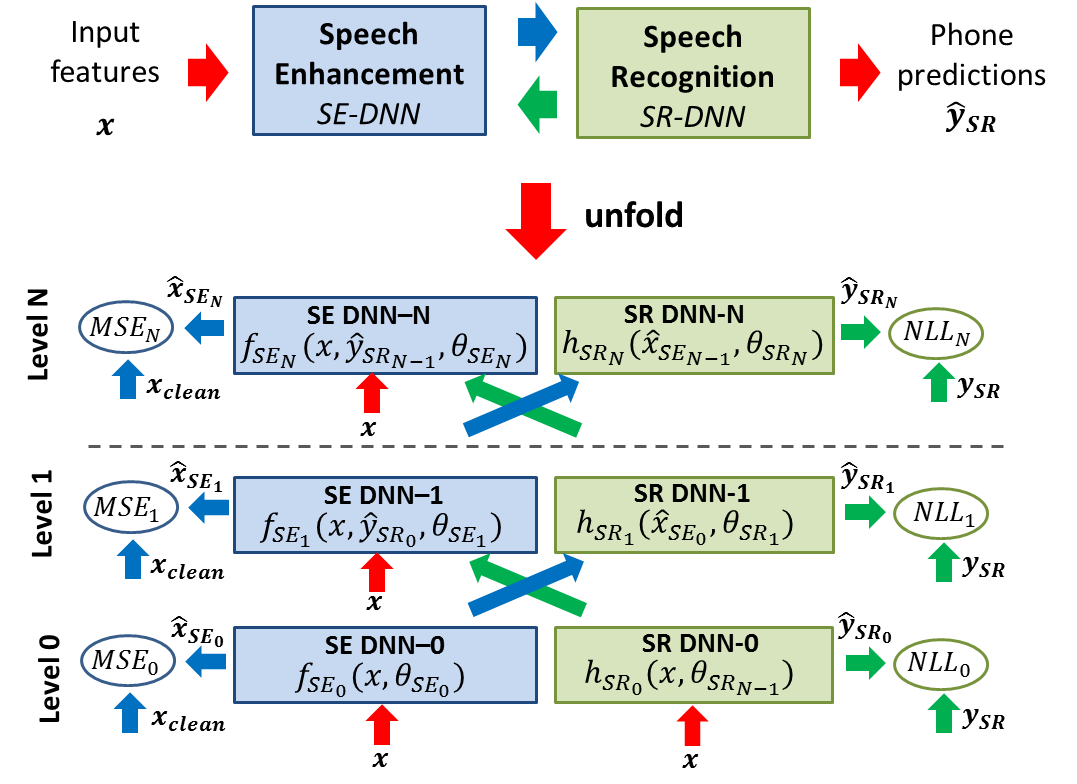}
\caption{The proposed network of deep neural networks.}
\label{fig:arch}
\end{figure}

The proposed network of deep neural networks is depicted in Fig.~\ref{fig:arch}. The upper part of the figure highlights the full communication between the speech enhancement and speech recognition DNNs. However, such a communication modality inherently entails a chicken-and-egg problem. This is caused by the fact that the speech recognizer is fed by the speech enhancement, which requires, in turn, also the speech recognition output itself to generate the enhanced speech. We can circumvent this issue by unfolding the proposed architecture to an arbitrary number of levels $L$, as shown in  Fig. ~\ref{fig:arch}. The resulting computational graph is built by concatenating several speech enhancement and speech recognition DNNs, which form various interaction levels. At the first level ($\ell=0$)  the two networks are independent, but a full communication is  established and continuously refined by progressively adding more levels to the architecture. 

An effective communication, however, can be impaired by the high dimensionality of the speech recognizer output $\hat{y}_{SR_\ell}$. This dimensionality derives from the number of considered context-dependent states, which typically ranges from 1000 to 4000 (depending on the phonetic decision tree and on the dataset). To avoid feeding the speech enhancement with such a high dimensional input, we jointly estimate the monophone targets (which are only some dozens). Similarly to \cite{multitask}, this is realized by adding an additional softmax classifier on the top of the last hidden layer of each speech recognition DNN.

The proposed architecture is jointly trained with the \textit{back-propagation through network} algorithm described in Alg.~\ref{alg}, where $x$ are the input features, $L$ the number of levels, $N$ the number of samples in the minibatch, $g$ the gradients and $\theta$ the DNN parameters. This algorithm is repeated for all the minibatches and iterated for several epoch until convergence.
The basic idea is to perform a forward pass, compute the loss functions at the output of each DNN (mean-squared error MSE for speech enhancement and negative multinomial log-likelihood NLL for speech recognition), compute the corresponding gradients, and back-propagate them.  In particular, following an approach inherited by recent joint training works, the gradient of each DNN is back-propagated through all the connected lower-level DNNs. Therefore, the speech enhancement parameter updates not only depend on the speech enhancement cost, but also on the higher-level speech recognition loss (line 11 of Alg. \ref{alg}). In this way the enhancement process is in part guided by the speech recognition cost function and it would hopefully be  able to provide enhanced speech which is more suitable for the subsequent speech recognition task. 
Similarly, the updates of the $\ell$-level speech recognizer also depend on the $\ell+1$ speech enhancement cost function. 
In general, as we discussed in \cite{ravanelli_SLT}, the integration of different gradients coming from the higher levels produces a regularization effect, which could significantly help the training of the system. According to this vision, the parameter $\lambda$, which weights the gradient coming from the higher levels, can be regarded as a regularization hyperparameter, whose optimal value can be determined on the development-set.

\begin{algorithm}[t!]
\caption{back-propagation through network algorithm}
\label{alg}
\begin{algorithmic}[1]
 \State \textbf{Forward Pass:} 
 \State Starting from input $x$ do a forward pass through the DNNs.
 
  \State \textbf{Compute Cost Functions ($\ell \in 0,..,L$):} 
  \State $MSE_{\ell}=\frac{1}{N}\sum_{n=1}^{N}(\hat{x}_{SE_\ell}-x_{clean})^2$
  \State $NLL_{\ell}=-\frac{1}{N}\sum_{n=1}^{N}y_{SR} log(\hat{y}_{SR_\ell})$ 
  
  \State \textbf{Gradient Computation ($\ell \in 0,..,L$):}
  \State $g_{SE_\ell}=\frac{\partial MSE_{\ell}}{\partial \theta_{SE_\ell}}$, $g_{SR_\ell}=\frac{\partial NLL_{\ell}}{\partial \theta_{SR_\ell}}$
  \State Back-propagate $g_{SE_\ell}$, $g_{SR_\ell}$ through $\ell-1$ level:
   \State $g_{SE_{\ell}\rightarrow SR_{\ell-1}}=\frac{\partial MSE_{\ell}}{\partial \theta_{SR_{\ell-1}}}$, $g_{SR_{\ell}\rightarrow SE_{\ell-1}}=\frac{\partial NLL_{\ell}}{\partial \theta_{SE_{\ell-1}}}$

  \State \textbf{Parameter Updates ($\ell \in 0,..,L$):}
   \State  $\theta_{SE_\ell} \gets \theta_{SE_\ell} - \eta  [(1-\lambda) g_{SE_\ell}+\lambda g_{SR_{\ell+1}\rightarrow SE_{\ell}}]$
   \State  $\theta_{SR_\ell} \gets \theta_{SR_\ell} - \eta  [(1-\lambda) g_{SR_\ell}+\lambda g_{SE_{\ell+1}\rightarrow SR_{\ell}}]$

\end{algorithmic}
\end{algorithm}

The joint training of the network of DNNs, however, can be complicated by the fact that the output distribution of the speech enhancement and speech recognition systems may change substantially during the optimization procedure. The speech recognition and speech enhancement modules would have to deal with an input distribution that is non-stationary and unnormalized. To mitigate this issue, we suggest to couple the proposed architecture with batch normalization. Batch normalization \cite{batchnorm}, which has been recently proposed in the machine learning community, addresses this concern (known as \textit{internal covariate shift}) by  normalizing the mean and the variance of each layer for each training mini-batch, and back-propagating through the normalization step. Similarly to what we observed in \cite{ravanelli_SLT}, batch normalization resulted crucial to significantly achieve better performance, to improve convergence of the proposed training algorithm, and to avoid any time-consuming pre-training steps.

\section{Related work}
The idea of using more than one neural network in the speech recognition process has long been explored. Examples of multi-DNN systems were, for instance, the so-called hierarchical bottleneck DNNs \cite{IEEEexample:hbn2,IEEEexample:bn6,tb}, which considered a cascade between a short-term and a long-term DNN to embedding longer-term information in the speech recognition process. 

More recently, the joint training methods outlined in Sec.\ref{sec:intro} have gained considerable attention. This work can be considered as an evolution of such approaches, in which  we employ a more advanced architecture based on a full communication between the DNNs. Similarly to this work, an iterative pipeline based on feeding the speech recognition output into a speech enhancement DNN has recently been proposed in \cite{joint6,ndnn1}. The main difference with our approach is that the later circumvent the chicken-and-egg problem by simply feeding the speech enhancement with the speech recognition alignments generated at the previous iteration, while our solution faces this issue by adopting the  unrolling procedure over different interaction levels previously discussed.

Our paradigm has also some similarities with traditional multi-tasking techniques \cite{multi_ow}. The main difference is that the latter are based on sharing some hidden layers across the tasks, while our method relies on  exchanging DNN outputs at various interaction levels. 
Finally, the proposed training algorithm has some aspects in common with the back-propagation through structure originally proposed for the parsing problem \cite{goller}.  The main difference is that the latter back-propagates the gradient through a tree structure, while the proposed variation back-propagates it on a less constrained network of components. Another difference is that in the original algorithm the same neural network is used across all the levels of the tree, while in this work different types of DNNs (i.e., speech enhancement and speech recognition) are involved. 

\section{Experimental Setup}

\subsection{Corpora and tasks}
\label{sec:corpora}
To provide an accurate evaluation of the proposed technique, the experimental validation has been conducted using different training datasets, different tasks and various environmental conditions. 
In particular, a set of experiments with TIMIT has been performed to test the proposed paradigm in low-resources conditions. To validate on a more realistic task, the proposed technique has also been evaluated on a WSJ task\footnote{The DIRHA-English dataset will be publicly distributed through the Linguistic Data Consortium (LDC). The scripts to generate the contaminated version of TIMIT will be available at \url{https://github.com/mravanelli}}.

The experiments with TIMIT are based on a phoneme recognition task (aligned with the Kaldi s5 recipe). The original training dataset has been contaminated with a set of impulse responses measured in a real apartment. The reverberation time ($T_{60}$) of the considered room is about 0.7 seconds. Development and test data have been simulated with the same approach, but considering a different set of impulse responses. 

The WSJ experiments are based on the popular wsj5k task (aligned with the CHiME 3 \cite{chime3} task) and are conducted under two different acoustic conditions. For the \textit{WSJ rev} case, the training set is contaminated with the same set of impulse responses adopted for TIMIT. For the \textit{WSJ rev+noise} case, we also added non-stationary noises recorded in a domestic context (the average SNR is about 10 dB). The test phase is carried out with the DIRHA-English corpus (real-data part), consisting of 409 WSJ sentences uttered by six native American speakers in the above mentioned apartment. More details on this corpus and on the impulse responses adopted in this work can be found in \cite{dirha_asru,rav_is16}.  

\subsection{System details}
\label{sec:details}
The features considered in this work are standard 39 Mel-Cepstral Coefficients (MFCCs) computed every 10 ms with a frame length of 25 ms. The speech enhancement DNNs are fed with a context of 21 consecutive frames and predict (every 10 ms) 11 consecutive frames of enhanced MFCC features.
The speech recognition DNNs are fed by such 11 speech enhanced frames and predict both context-dependent and monophone targets at their output. 
All the layers used Rectified Linear Units (ReLU), except for the output of the speech enhancement DNNs (linear) and the output of  the speech recognition modules (softmax).
Batch normalization \cite{batchnorm} and dropout \cite{dropout} are employed for all the hidden layers. 
The labels for the speech enhancement DNN (denoted as $x_{clean}$ in Fig. ~\ref{fig:arch}) are the MFCC features of the original clean datasets.
The labels for the speech recognition DNN (denoted as $y_{SR}$ in Fig. ~\ref{fig:arch}) are derived by performing a forced alignment procedure on the original training datasets. See the standard s5 recipe of Kaldi for more details \cite{kaldi_short}.

The weights of the networks are initialized according to the \textit{Glorot} initialization \cite{xavier}, while biases are initialized to zero.
Training is based on a standard Stochastic Gradient Descend (SGD) optimization with mini-batches $N$ of size 128. The performance on the development set is monitored after each epoch and the learning rate $\eta$ is halved when the performance improvement is below a certain threshold. The training ends when no significant improvements have been observed for more than four consecutive epochs. 

The main hyperparameters of the system (i.e., learning rate $\eta$, number of hidden layers, hidden neurons per layer, dropout factor, gradient weighting factor $\lambda$ and number of unfolding levels $L$) have been optimized on the development set. As a result, speech enhancement and speech recognition DNNs with 4 hidden layers of 1024 neurons and DNNs with 6 hidden layers of 2048 neurons are employed for TIMIT and WSJ tasks, respectively. The initial learning rate is 0.08, the dropout factor is 0.2 and the considered number of levels is 3 ($l=0,..,2$).  Similarly to \cite{ravanelli_SLT}, $\lambda$ is fixed to 0.1.

The proposed system, which has been implemented with Theano \cite{theano}, 
has been coupled with the Kaldi toolkit \cite{kaldi_short} to form a context-dependent DNN-HMM speech recognizer.

\section{Results}
\subsection{Close-talking baselines}
\label{sec:ct_baseline}
The Phoneme Error Rate (PER\%) obtained by decoding the original test sentences of TIMIT is $19.5\%$ (using DNN models trained with the original dataset). The Word Error Rate (WER\%) obtained by decoding the close-talking DIRHA-English WSJ sentences is $3.3\%$.  It is worth noting that, under such favorable acoustic conditions, the DNN model leads to a very accurate sentence transcription, especially when coupled with a language model.

\subsection{Network of DNNs performance}
\label{sec:jt_pers}
The proposed network of DNNs approach is compared in Table \ref{tab:res1} with other competitive systems. 
The first line reports the results obtained with a single neural network. In this case, only the speech recognition labels are used and the DNN is not forced to perform any speech enhancement task.
The second line shows the performance obtained when the single DNN is coupled with a traditional multi-task learning, in which a speech enhancement and a speech recognition task are simultaneously considered. This multi-task architecture shares the first half of the hidden layers across the tasks, while the second half of the architecture is task-dependent. This approach aims to discover (within the shared layers) more general and robust features which can be exploited to better solve both correlated tasks.  
The third line reports the performance achieved with the joint training approach recently proposed in \cite{ravanelli_SLT}. In this case a bigger DNN composed of a cascade of a speech enhancement and a speech recognition DNNs is jointly trained by back-propagating the speech recognition gradient also into the speech enhancement DNN. 
The last line finally shows the performance achieved with the proposed network of deep neural network approach. To allow a fair comparison, batch normalization is adopted for all the considered systems.

Table \ref{tab:res1} highlights that the proposed approach significantly outperforms all the single DNN systems. For instance, a relative improvement of about 14\%  over the single DNN baseline is obtained for the \textit{WSJ rev+noise} case. The network of deep neural networks also outperforms the considered joint training method. This result suggests that the improved cooperation between the networks achieved with our full communication scheme can overtake the standard DSR pipeline based on a partial and unidirectional information flow (which is still considered  in  the context of joint training approaches).     

Table \ref{tab:res2} shows the results obtained by decoding the speech recognition output at the various levels of the proposed architecture (denoted as $\hat{y}_{SR_0}$, $\hat{y}_{SR_1}$,  $\hat{y}_{SR_2}$ in Fig. \ref{fig:arch}). One can note that the performance become progressively better as the level of the network of DNNs increases. As expected, the first level speech recognizer (\textit{SR DNN-0}) performs similarly to the single DNN baseline. The second level (\textit{SR DNN-1}) is based on a simple cascade between a speech enhancement and a speech recognition DNNs,  and thus provides results similar to that obtained with standard joint training. The third level (\textit{SR DNN-2}) achieves the best performance,  confirming that the progressive interaction of the DNNs involved in the DSR process helps in improving the system performance. No additional benefits have been observed for the considered tasks by adding more than 3 levels.
\begin{table}[t!]
\centering
\tabcolsep=0.28cm
    \begin{tabular}{ | l | c | c | c | c | }
    \cline{1-4}
    \multirow{2}{*}{\backslashbox{\em{System}}{\em{Dataset}}} & \multicolumn{1}{ | c |}{TIMIT}  & \multicolumn{1}{ | c |}{WSJ} & \multicolumn{1}{ | c |}{WSJ}  \\ \cline{2-4}
    & \textit{rev} & \textit{rev} & \textit{rev+noise}  \\ \hline
      Single DNN & 31.9  & 8.1 & 14.3    \\ \hline
      Single DNN +multitask & 31.4  & 8.1 & 13.8    \\ \hline
      Joint SE-SR training & 29.1  & 7.8 & 12.7    \\ \hline
      Network of DNNs & \textbf{28.7}  & \textbf{7.6} &  \textbf{12.3}    \\ \hline  
    \end{tabular}
\caption{Performance of the proposed network of DNN approach compared with other competitive DNN-based systems (PER\% for TIMIT, WER\% for WSJ).}
\label{tab:res1}
\end{table}
\label{sec:bn_exp}

\begin{table}[t!]
\centering
\tabcolsep=0.28cm
    \begin{tabular}{ | l | c | c | c | c | }
    \cline{1-4}
      Dataset & Level 0  & Level 1 & Level 2    \\ \hline
      TIMIT rev & 31.4 & 29.1 & 28.7  \\ \hline
      WSJ rev & 8.0 & 7.7 & 7.6  \\ \hline
      WSJ rev+noise & 14.3 & 12.7 & 12.3  \\ \hline 
    \end{tabular}
\caption{Performance of the proposed network of DNN achieved at various levels of the architecture.}
\label{tab:res2}
\end{table}
\label{sec:bn_exp}


\subsection{Alternative architectures}
\begin{figure}[t!]
\centering
\includegraphics[width=0.20\textwidth]{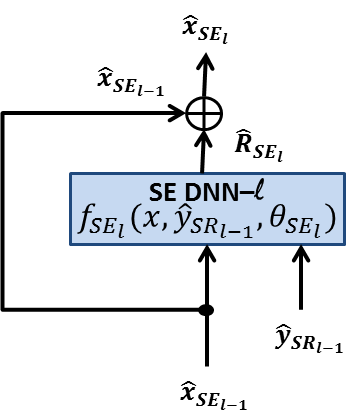}
\caption{The architectural variation inspired by residual networks.}
\label{fig:arch2}
\end{figure}

The architecture depicted in Fig. ~\ref{fig:arch} represents only one of the possible ways to implement the proposed network of DNN paradigm. An alternative solution would be to share the parameters across the various speech enhancement and speech recognition DNNs. However, results (not reported here) show no benefits from this approach. Another possible modification is represented by pre-training each DNN before performing the back-propagation through network. As observed in \cite{ravanelli_SLT}, when batch normalization is adopted, no benefits from any pre-training methods have been observed. 
Although a more detailed exploration of alternative solutions is under study, in this section we report some preliminary results obtained by evolving the proposed method with an architecture partly-inspired by residual networks (ResNets) \cite{res_net}. ResNets have recently achieved state-of-the-art performance in computer vision and are based on the idea that, instead of learning a distribution directly, we can learn more easily the residual functions with reference to the input layer.  In this work, as shown in Fig. \ref{fig:arch2} and Eq.\ref{reseq}, we employ a ResNet-inspired architecture for the speech enhancement DNNs:
\begin{equation}  
\label{reseq}
\hat{R}_{SE_l}=\hat{x}_{SE_{l-1}}-\hat{x}_{SE_{l}}\ .
\end{equation}
This choice is motivated by the fact that we progressively expect less variation between the input and the output of the speech enhancement DNN as the levels increase. Directly learning the residual can thus make the training of the higher levels easier.
The results achieved with the ResNet inspired architecture are reported in Tab. \ref{tab:res4} and confirm the effectiveness of this approach, showing a coherent improvement over all the considered tasks.

\begin{table}[t!]
\centering
\tabcolsep=0.20cm
    \begin{tabular}{ | l | c | c | c | c | }
    \cline{1-4}
    \multirow{2}{*}{\backslashbox{\em{System}}{\em{Dataset}}} & \multicolumn{1}{ | c |}{TIMIT}  & \multicolumn{1}{ | c |}{WSJ} & \multicolumn{1}{ | c |}{WSJ}  \\ \cline{2-4}
    & \textit{rev} & \textit{rev} & \textit{rev+noise}  \\ \hline
      Original Arch. & 28.7  & 7.6 & 12.3    \\ \hline
      ResNet-inspired & \textbf{28.4}  & \textbf{7.5} & \textbf{12.0}    \\ \hline
    \end{tabular}
\caption{Performance with the ResNet-inspired architecture.}
\label{tab:res4}
\end{table}

\section{Conclusion and Future Works}
In this paper, we proposed a novel architecture for distant speech recognition based on a network of deep neural networks. The experiments, conducted considering different tasks, datasets and acoustic conditions, show that our method is effectively able to take advantage of a full communication between a speech enhancement and a speech recognizer, leading to a performance significantly better than that obtained with more standard DSR pipelines.

This work, however, represents only a first step towards more advanced architectures able to better cooperate and communicate each other to achieve a common goal. Our future research efforts will be thus focused on improving the current paradigm. In this paper, for instance, we only considered speech enhancement and speech recognition DNNs. Nevertheless, the proposed approach is a general framework that can be extended in a straightforward way by involving other modules, eventually including DNNs for acoustic scene classification, voice activity detection and speaker identification. Moreover, a more detailed exploration of alternative architectures and the natural extension of this paradigm to RNNs will also be considered as future research.


\bibliographystyle{IEEEbib}
\bibliography{strings,refs}

\end{document}